\documentclass[11pt]{article}
\usepackage{microtype}
\usepackage{epstopdf}
\usepackage{amsmath,amssymb}
\usepackage{amsfonts,bm,bbm} 
\usepackage{amssymb,amsthm,enumitem}

\usepackage{multirow}
\usepackage{xcolor,graphicx,float,booktabs}
\usepackage{tabularx}
\newcolumntype{L}[1]{>{\raggedright\arraybackslash}m{#1}}
\newcolumntype{C}[1]{>{\centering\arraybackslash}m{#1}}
\newcolumntype{R}[1]{>{\raggedleft\arraybackslash}m{#1}}
\usepackage{tikz}
\usetikzlibrary{calc,arrows.meta,positioning,decorations.pathmorphing,shadows.blur}
\usepackage{placeins}
\usepackage[most]{tcolorbox}
\usepackage{listings}
\usepackage{inconsolata} 

\usepackage{epstopdf}

\usepackage{verbatim} 
\allowdisplaybreaks[4]
\usepackage{authblk}
\usepackage{cite}
\usepackage{hyperref}
\hypersetup{hypertex=true,
colorlinks=true,
linkcolor=blue,
anchorcolor=blue,
citecolor=blue}

\usepackage[round]{natbib}
\usepackage[
margin=1in,
includefoot,
footskip=30pt,
]{geometry}

\usepackage[algo2e,ruled,vlined]{algorithm2e}

\numberwithin{equation}{section}

\RequirePackage{endnotes}

\title{Mechanism-Faithful Queueing Simulation Model Translation with Large Language Model Support}

\author[1]{Jun-Qi Chen\footnote{jqchen@ruc.edu.cn}}
\author[2]{Kun Zhang*\footnote{ kunzhang@ruc.edu.cn}}
\author[1]{Rui Zheng\footnote{ zr20230000789@ruc.edu.cn}}
\author[3]{Ying Zhong\footnote{yzhong4@uestc.edu.cn}}

\affil[1]{Institute of Statistics and Big Data, Renmin University of China, No. 59 Zhongguancun Street, Haidian District, 100872, Beijing, China}
\affil[2]{School of Information, Renmin University of China, No. 59 Zhongguancun Street, Haidian District, 100872, Beijing, China}
\affil[3]{School of Management and Economics, University of Electronic Science and Technology of China, No.2006, Xiyuan Ave, West Hi-Tech Zone, 611731, Chengdu, China}

\begin{document}
\normalsize

%
%
%
%
%
%
%

\maketitle

\begin{abstract}
Queueing simulation studies often require substantial manual effort to translate conceptual system descriptions into executable programs and to verify that the implemented mechanisms match the intended queueing logic. Although large language models (LLMs) may produce executable scripts, executability alone is insufficient when arrival, routing, interruption, or reporting logic is wrong. This study presents a simulation-oriented support framework for \texttt{SimPy}-based queueing model translation. We propose a category-template framework for mechanism coverage with a staged adaptation workflow that targets structured event logic and common simulation-specific failure modes. On held-out task instances, the adapted models improve executability, output-format compliance, and instruction-mechanism consistency across basic, behavioral, and networked queueing settings, so the generated scripts are more reliable as queueing simulation scripts. Error analysis shows better preservation of routing semantics and interruption-resume logic, while also exposing remaining weaknesses in multi-node transfer and residual-service updates. Overall, the results suggest that the proposed framework can act as a simulation-faithful generator for more standardized and reproducible queueing model construction.

\end{abstract}

\section{Introduction}

Discrete-event simulation (DES) is a foundational analytical tool in operations research and management science, enabling researchers and practitioners to study stochastic systems through explicit modeling of events, resources, and state transitions. Among various DES applications, queueing-system simulation plays a central role in analyzing customer waiting behavior, resource utilization, congestion patterns, and overall service efficiency. Its practical importance has been demonstrated across diverse domains, including healthcare operations \citep{Dong2020OR}, 
manufacturing systems \citep{Huang2022NRL}, and service operations \citep{Cezik2008CallCenter}. Across these settings, queueing simulation has served as a valuable tool for evaluating operational policies and improving system performance.

As \cite{Banks2013DiscreteEvent} note, a rigorous simulation study in operations research and management science typically follows a structured workflow consisting of problem formulation and model conceptualization, model translation, verification and validation, experimentation, and reporting. After defining the decision context and research objectives, researchers abstract the essential system mechanisms into a conceptual model and collect the required input data. This conceptual model must then be translated into an executable simulation program, which is subsequently verified and validated to ensure logical correctness and fidelity to the real system. Once validated, the model supports experimental design, production runs, and statistical analysis, with findings documented for interpretation and potential implementation. Within this workflow, the translation from conceptual model to executable program is a critical step, because it connects abstract system understanding to the simulation artifact used for analysis and decision support.

Within this workflow, one of the most labor-intensive parts is the translation of a conceptual model into an executable simulation program, together with the subsequent verification and validation needed to ensure that the resulting model is usable for analysis. Specifically, researchers must (i) translate the conceptual system description into executable simulation code by implementing arrival processes, service mechanisms, event scheduling, and state-update rules; (ii) repeatedly run, verify, and debug the program to ensure that the implemented logic is internally consistent and free of coding errors; and (iii) validate the simulation outputs against theoretical results, approximation formulas, or empirical benchmarks, followed by iterative code refinement when discrepancies are detected.

For instance, even in a common multi-server queueing system, translating a conceptual description into correct and reliable simulation code is often iterative and time-consuming. Researchers typically begin with a baseline implementation that encodes the arrival process, service mechanisms, and queueing discipline, and then obtain preliminary estimates of key performance measures. When these outputs deviate from theoretical results or empirical benchmarks, extensive diagnostic effort is required to identify the sources of discrepancy. This diagnosis commonly involves repeatedly inspecting random number generation, event-scheduling logic, state-update rules, and performance-measure collection (e.g., warm-up handling), followed by incremental code modification and re-verification. As models become more complex through networked queues, routing, heterogeneous servers, or state-dependent behavior, coupling among code components intensifies, which makes debugging and validation increasingly reliant on manual expertise and trial-and-error. As a result, model translation and refinement become a major bottleneck in simulation studies.

Recent advances in large language models (LLMs) have created new opportunities to alleviate this bottleneck. LLMs have been increasingly studied for code generation across a variety of programming tasks \citep{Jiang2026ACM}. In the simulation domain, the overview and perspective article by \cite{HongNelson2026EJOR} suggests that ``{\it At a basic level, language-to-simulation modeling could become feasible. ..., users might describe the model in natural language, which LLMs could then translate into executable simulation code.}" Motivated by this vision, we aim to develop an LLM-assisted method for simulation model translation, with a particular focus on queueing systems, so that domain experts can describe target queueing systems in natural language and obtain the corresponding simulation code. Beyond mere executability, our method is designed to support trustworthy conceptual-to-executable translation while preserving mechanism fidelity.

However, directly applying existing general-purpose LLMs to queueing-system simulation may reveal a fundamental mismatch between executability and mechanism fidelity. In operations research settings, the goal is to translate queueing concepts into executable models while preserving the intended queueing mechanisms. A generated Python script may run without error and still be behaviorally invalid if it misrepresents the queueing discipline, routing logic, interruption handling, or the definition and collection of queueing KPIs such as waiting time, utilization, and abandonment. What is missing is not another code generator, but a method for conceptual-to-executable translation that preserves mechanism fidelity.

A key obstacle to applying LLMs to SimPy-based queueing-system simulation is the scarcity and uneven quality of suitable training data. This obstacle is reflected in at least three ways. First, only a limited amount of simulation code is publicly available, even when examples from official SimPy tutorials and online repositories are taken into account. Second, the available code examples cover only a narrow range of scenarios and do not adequately represent the diversity of queueing systems encountered in practice. Third, only a small portion of the available code follows a unified and standardized output format. As \citet{Huang2025ORLM} show, one possible response is to leverage LLMs to automatically generate simulation code. However, such generated code may still contain logical errors in queueing mechanisms or exhibit unstable execution behavior, and therefore requires further validation and revision. To address these issues, we propose a simulation-oriented data construction and adaptation workflow with three stages: the first two stages use supervised fine-tuning (SFT) to address data scarcity and uneven data quality, and the third stage applies Direct Preference Optimization (DPO) \citep{rafailov2023DPO} to better preserve queueing semantics.

The three stages are designed to address different simulation needs in queueing model translation. In \textbf{Stage~I}, we construct a large-scale training dataset with broad coverage of queueing-system scenarios using a \emph{Category-Template Framework (CTF)}.  This stage addresses the simulation need for explicit mechanism coverage by organizing diverse natural-language descriptions and multiple \texttt{SimPy} implementation styles under a standardized translation scheme. In \textbf{Stage~II}, we strengthen the model's understanding of queueing mechanisms through supervised fine-tuning with masked code completion. This stage addresses the simulation need to reconstruct structured event logic by training the model to complete functional code segments that must remain consistent with the surrounding process interactions. In \textbf{Stage~III}, we perform DPO-based preference optimization. This stage addresses recurrent simulation-specific failure modes by contrasting preferred code scripts that faithfully reflect the intended queueing mechanism with rejected ones that may run successfully but contain mechanism-level logic errors.

Evaluations on a 600-task benchmark demonstrate clear and consistent gains in executability rate (EXE), output-format compliance (FMT), and instruction-mechanism consistency (IMC). These gains are important because, from a simulation perspective, the objective is not generic code generation quality, but more reliable conceptual-to-executable translation while preserving mechanism fidelity and reproducibility in queueing studies. In particular, we study whether the resulting models can generate simulation-faithful scripts whose implemented mechanisms and KPI behavior remain consistent with the intended queueing logic. Detailed quantitative results are reported in Section~\ref{subsec:main-results}.

Our contributions are as follows:

\begin{enumerate}
	
	\item {\it Systematic bridging of conceptual and executable queueing models:} We propose CTF, a structured framework that maps queueing mechanisms (e.g., stochastic arrivals, preemption, routing) to mechanism-grounded \texttt{SimPy} implementations. This framework provides a reproducible foundation for translating DES conceptual models into executable programs while reducing ad hoc variation in mechanism implementation.
	
	\item {\it Mechanism-oriented learning for queueing model translation:} We develop a staged adaptation workflow that improves the preservation of queueing mechanisms beyond syntactic executability, especially for event scheduling, routing logic, and interruption handling.
	
	\item {\it Practical support for queueing simulation workflows:} Across diverse queueing scenarios, compact locally deployable models achieve stronger EXE/FMT/IMC performance and can serve as simulation-faithful script generators for more standardized and reproducible queueing model construction.

\end{enumerate}

\section{Related Work}

Our work lies at the intersection of several related literature streams, including the modeling bottleneck in discrete-event simulation, recent advances in LLM-based code generation, and emerging applications of LLMs in operations research. These streams together motivate the need for a mechanism-faithful approach to queueing simulation model translation, while also highlighting the gap between generic code generation and faithful simulation modeling. We review the most relevant studies below and position our contribution within this intersection.

\subsection{The Modeling Bottleneck in Queueing Simulation}

Queueing systems are among the most classical and extensively studied application areas of DES. Although the theoretical foundations of queueing simulation are well established \citep{Banks2013DiscreteEvent}, developing and implementing queueing simulation models for complex real-world settings, such as healthcare patient flow \citep{Dong2020OR}, multiskill call centers \citep{Cezik2008CallCenter}, and manufacturing systems \citep{Huang2022NRL}, remains an expertise-dependent process.

To improve transparency and reproducibility, implementing simulation models in code-based frameworks such as \texttt{SimPy} has become increasingly popular \citep{Mohapatra2025simpy, zinoviev2024simpy}. However, translating a conceptual system description into an executable simulation program remains labor-intensive. Based on an empirical study of expert DES modelers, \citet{Tako2011Codeh} report that analysts spend considerable time on model coding and debugging. This model translation phase, which requires arrivals, services, and resource constraints to be encoded correctly in executable form, represents a major bottleneck in queueing simulation studies.

\subsection{LLMs for Automated Code Generation}

Recent advances in LLMs have led to growing interest in automated code generation. In the broader context of program synthesis, large-scale models such as AlphaCode \citep{li2022AlphaCode} and Code Llama \citep{roziere2024codellama} demonstrate that LLMs can generate executable programs and achieve strong performance on standardized coding benchmarks. Frameworks such as InterCode \citep{Yang2023InterCode} further emphasize executability and runtime behavior as central evaluation criteria, thereby reinforcing a benchmark-oriented view of code generation quality.

However, in the context of these general-purpose code generation tasks, the primary evaluation metrics such as pass@k focus predominantly on syntactic correctness, generic algorithmic logic, and the absence of runtime errors. This perspective, rooted in natural language processing (NLP) and software engineering, does not fully capture the rigorous requirements in operations research and simulation settings. These benchmarks do not directly assess whether generated code constitutes a valid simulation script for experimentation.

\subsection{LLMs in Operations Research and Simulation}

Building on these advances, recent studies have begun to explore the use of LLMs for modeling tasks in operations research and simulation. For example, \cite{Huang2025ORLM} propose ORLM, which fine-tunes open-source LLMs on domain-specific synthetic data to generate optimization formulations and solver code. Related work has also begun to examine LLM-based support for simulation modeling: \cite{Gao2024LLM} survey LLM-empowered agent-based modeling and discuss challenges related to alignment and action generation.

Despite these explorations, the automation of executable and mechanism-faithful queueing-system simulation models remains largely unexplored. The critical gap lies in the distinction between generic code correctness and mechanism fidelity. For a general software task, a syntactically valid and executable script is often deemed a success. However, in queueing-system simulation, a generated Python script may run perfectly without any runtime errors, yet completely misrepresent the underlying queueing discipline, resource preemption rules, or event-scheduling logic. In operations research settings, such hidden mechanistic errors render the simulation entirely invalid for decision-making.

Therefore, our work is distinctly driven by a {\it mechanism-consistency orientation}. Rather than merely maximizing the executability of Python scripts, our workflow is designed to preserve arrival, service, routing, interruption, preemption, and KPI logic that determine whether a queueing model can be used reliably for experimentation and decision support. Existing work either addresses general code generation or operations research formulation and solver modeling; by contrast, this study focuses on the faithful construction of executable queueing simulation models. Table~\ref{tab:rw-positioning} summarizes this positioning relative to the related literature streams.

\begin{table}[htbp]
	\centering
	\setlength{\tabcolsep}{3pt}
	\renewcommand{\arraystretch}{0.9}
	\begin{tabular}{lccc}
		\toprule
		\multirow{2}{*}{Work Type}	& Mechanism  & Multi-stage  & Common Error \\
		& Consistency & Training &  Analysis \\
		\midrule
		General code LLM benchmarks & Partial & No & No \\
		OR optimization-code generation & Partial & Partial & No \\
		LLM-assisted simulation surveys & Conceptual & No & No \\
		This work & Yes & Yes & Yes \\
		\bottomrule
	\end{tabular}
	\caption{Positioning of this study relative to adjacent literature streams. (In this table,  ``Yes'' indicates that a literature stream addresses the corresponding dimension explicitly and systematically; ``No'' indicates that it does not; ``Partial'' indicates limited or incomplete coverage; and ``Conceptual'' indicates discussion at the conceptual level without a concrete method or evaluation design.)}
	\label{tab:rw-positioning}
\end{table}

\section{Problem Formulation and Evaluation Dimensions}\label{sec:problem}

We formalize queueing simulation code generation as a mapping from a natural-language simulation specification to an executable \texttt{SimPy} program:
\begin{align*}
	\hat y=f_{\theta}(x),\quad x \in \mathcal{X},\ \hat y \in \mathcal{Y},
\end{align*}
where $x$ specifies mechanisms (arrival/service rules, queue discipline, routing, interruption, reporting constraints), and $\hat y$ is generated Python code expected to run and represent the required queueing semantics. The input is therefore a queueing system specification or conceptual description, and the output is an executable queueing model with valid simulation logic. The object of interest is not code text per se, but an executable simulation script that can be used in experimentation and decision support.

For each generated script $\hat y$, we evaluate three dimensions:
\begin{enumerate}
	\item Executability (EXE): 1 if the script runs to completion within timeout and without runtime errors; otherwise 0.
	
	\item Output-format compliance (FMT): 1 if the output strictly satisfies the required reporting contract (labels and numeric precision); otherwise 0.
	
	\item Instruction-mechanism consistency (IMC): 1 if the implemented queueing mechanisms match instruction semantics; otherwise 0.
\end{enumerate}
These metrics are intended to assess deployability, reporting readiness, and mechanism fidelity, respectively. Among the three metrics, IMC is the closest proxy to simulation validity because it evaluates whether the generated code preserves the instructed operational mechanisms. Let $m \in$ \{EXE, FMT, IMC\} denote one evaluation metric, and let $m(x,\hat y)\in\{0,1\}$ be the corresponding indicator for whether generated script $\hat y$ satisfies that metric on input $x$. For a test set $\mathcal{D}_{\text{test}}$, pass rates are reported as percentages:
\begin{align*}
	\mathrm{Rate}_{m}(f_{\theta})=\frac{1}{|\mathcal{D}_{\text{test}}|}\sum_{x\in \mathcal{D}_{\text{test}}} \mathbf{1}\{m(x,\hat y)=1\} \times 100.
\end{align*}

To complement the pass/fail metrics with mechanism-level insight, we analyze recurrent implementation mistakes in Section~\ref{subsec:EA}. This section focuses on common error patterns and their impact on queueing semantics and KPI reliability. A script that passes EXE/FMT but fails IMC should therefore be viewed as deployable code, but not as a trustworthy simulation model.

To avoid leakage, all stages enforce instruction-level disjointness among training construction pools, Stage-III query data, and the final test set.

\section{Simulation-Oriented Multi-Stage Framework}

The proposed framework consists of three stages and has a single objective: to support standardized conceptual-to-executable queueing model translation while improving mechanism fidelity in generated scripts. Stage~I builds a broad validated instruction--code corpus under CTF for foundational SFT. Stage~II adds masked completion to strengthen mechanism reconstruction. Stage~III applies preference optimization to suppress recurrent semantic failures in full-script generation. These mechanisms affect simulation behavior in distinct ways: routing affects flow balance and throughput, interruption handling affects waiting time and service completion, and preemption changes service order and downstream KPI behavior.

\subsection{Stage I: Instruction-Code Data Construction and Initial Supervised Fine-Tuning}

\subsubsection{Motivation of Stage I}

In Stage~I, the simulation goal is to establish mechanism coverage and a standardized translation framework: the model should learn to generate executable \texttt{SimPy} scripts that preserve queueing logic and output required KPIs in the specified format. Achieving this goal requires training data with both mechanism coverage and representation diversity. We therefore propose CTF to systematically construct validated instruction--code pairs across heterogeneous queueing scenarios.

\subsubsection{Category-Template Framework}

CTF consists of three components:
\begin{enumerate}
	\item \textbf{Queueing category}: specifies the queueing scenario that represents a type of queueing task;
	\item \textbf{Instruction template}: provides alternative linguistic formulations describing the same queueing task for constructing the instruction;
	\item \textbf{Code template}: offers different code styles for generating the corresponding \texttt{SimPy} simulation code.
\end{enumerate}

\begin{table}[t]
	\centering
	\footnotesize
	\renewcommand{\arraystretch}{0.6} 
	\begin{tabular}{p{3.0cm} p{3.15cm} p{4.3cm} p{1.9cm}}
		\toprule
		\textbf{Category} & \textbf{Defining Mechanism} & \textbf{Represented Mechanisms} & \textbf{Complexity} \\
		
		\midrule
		\multicolumn{4}{l}{\textit{(A) Basic Systems}} \\
		\midrule
		finite capacity & Limited queue length $K$ & Blocking and system loss constraints & Simple \\
		general distributions & Non-exponential service or interarrival & Arbitrary stochastic variability & Simple \\
		multi server sched rules & Multiple servers with simple scheduling & Load balancing across identical servers & Simple  \\
		\midrule
		\multicolumn{4}{l}{\textit{(B) Intermediate Behavioral Extensions}} \\
		\midrule
		balking reneging & Customer balking and reneging & Queue avoidance and impatience behavior & Intermediate \\
		batch arrivals & Grouped Poisson arrivals & Burst-type demand or batch processing & Intermediate \\
		multi class customers & Distinct customer classes & Heterogeneous demand and service differentiation & Intermediate \\
		piecewise arrival & Time-varying or piecewise arrival rate & Stepwise nonstationary input process & Intermediate\\
		production kanban & Work-in-process control via tokens & Inventory-limited manufacturing systems & Intermediate \\
		\midrule
		\multicolumn{4}{l}{\textit{(C) Complex Networked or Multi-Mechanism Systems}} \\
		\midrule
		breakdown maintenance & Random server failures and repairs & Reliability and maintenance effects & Complex  \\
		parallel two resources & Joint resource coordination & Parallel processing or shared-resource synchronization & Complex  \\
		open network & Exogenous arrivals to multiple nodes & Distributed and interconnected queueing systems &Complex  \\
		feedback network & Internal routing between nodes & Reentrant service flow and job feedback & Complex  \\
		\bottomrule
	\end{tabular}
	\caption{Overview of queueing categories by increasing system complexity.}
	\label{tab:categories}
\end{table}

Each category represents a mechanism family; for each category, we maintain matched instruction and code templates. An instruction--code pair is generated by sampling one category, one instruction template, and one code template, then instantiating mechanism parameters. This design jointly controls mechanism coverage and linguistic/structural diversity, and it helps standardize model translation across queueing classes so that ad hoc implementation variability is reduced.

{\linespread{1.0}
	\begin{figure}[t]
		\small
		\centering
		\begin{tcolorbox}[enhanced, width=.96\columnwidth, colback=white, colframe=black!60,
			boxsep=1pt,left=1pt,right=1pt,top=1pt,bottom=1pt,
			title={\textbf{Instruction Templates for the \emph{Batch Arrivals} Category}}]
			
			\textbf{Prompt~1 (T0) --- Concise and instructional}\\[-1pt]
			\emph{Batch arrival scenario: external arrival rate $\lambda{=}0.641$, batch size $15$, and service rate $\mu{=}0.982$. 
				Run the simulation up to time $t{=}1100$. 
				Please implement an \textbf{executable} Python+SimPy script that runs until the specified end time and prints \textbf{exactly two lines (six decimal places)}: \textit{Average waiting time} and \textit{Utilization}. 
				Return \textbf{only} the raw Python source code---no explanations or markdown formatting.}\\[3pt]
			
			\textbf{Prompt~2 (T1) --- Explanatory and context-rich}\\[-1pt]
			\emph{Consider a single-server batch-arrival model where inter-batch arrivals follow an exponential distribution with rate $\lambda{=}0.641$, and each batch contains $15$ customers arriving simultaneously. 
				The service rate is $\mu{=}0.982$, and the simulation runs until $t{=}1100$. 
				Provide a \textbf{fully executable} SimPy script that prints two summary statistics at the end, \textit{Average waiting time} and \textit{Server utilization}, each formatted to six decimal places, without any additional text or explanations.}\\[3pt]
			
			\textbf{Prompt~3 (T2) --- Another variant}\\[-1pt]
			\emph{Simulate a single-server batch arrivals system where the external arrival rate is $\lambda{=}0.641$, each batch contains $15$ customers, and the service rate is $\mu{=}0.982$. 
				Run the simulation until time $t{=}1100$, and print exactly two lines (six decimals): \textit{Average waiting time} and \textit{Utilization}. 
				Only output the raw Python code without explanations or markdown.}
			
		\end{tcolorbox}
		\caption{Three instruction templates for the \emph{batch arrivals} category.}
		\label{fig:prompt-variants}
	\end{figure}
}

\textbf{Queueing Category. }  We define twelve queueing categories (Table~\ref{tab:categories}) to cover the mechanism space most relevant to practical DES modeling: arrival/service dynamics, server configuration, capacity limits, behavioral effects (balking/reneging), and network routing/feedback. For mechanism curriculum design, these categories are grouped into three difficulty levels: basic stationary systems, intermediate behavioral extensions, and complex multi-mechanism/networked systems. This structure is used to control coverage and difficulty during data construction, and to support category-level diagnosis in later experiments.

\textbf{Instruction Templates. } Instruction templates provide paraphrastic variants of the same simulation task so the model learns mechanism invariance under different phrasings. Representative variants are provided in Figure~\ref{fig:prompt-variants}.

\textbf{Code Templates. } Code templates provide three different styles for the same mechanism (procedural, object-oriented, and functional), which improves robustness to coding-style variation in deployment. Representative fragments and the full category-template library are provided in Figure~\ref{fig:template-variants}.

Overall, CTF provides explicit mechanism coverage and representation diversity while keeping generation rules reproducible. It standardizes model translation across queueing classes and improves reproducibility. As a reusable translation framework, CTF standardizes how recurring queueing mechanisms are turned into executable \texttt{SimPy} structures.

{\linespread{0.8}
	\begin{figure}[t]
		\small
		\centering
		\begin{tcolorbox}[enhanced, width=.94\columnwidth, colback=white, colframe=black!40,
			boxsep=1pt,left=1pt,right=1pt,top=1pt,bottom=1pt,
			title={\textbf{Implementation Templates for \emph{Batch Arrivals}}}]
			
			\textbf{T0 (Procedural)} --- flat script using global functions.\\ \vspace{-9pt}
			\begin{lstlisting}[language=Python]
				def customer(env, server, waits, busy, t_arr):
				with server.request() as req:
				yield req
				st = env.now
				yield env.timeout(random.expovariate(SERVICE_RATE))
				waits.append(st - t_arr)
			\end{lstlisting}
			
			\textbf{T1 (Object-oriented)} --- encapsulated as a simulation class.\\ \vspace{-9pt}
			\begin{lstlisting}[language=Python]
				class BatchSim:
				def customer(self, t_arr):
				with self.res.request() as req:
				yield req
				st = self.env.now
				yield self.env.timeout(random.expovariate(SERVICE_RATE))
				self.waits.append(st - t_arr)
			\end{lstlisting}
			
			\textbf{T2 (Functional)} --- modular structure with helper functions.\\ \vspace{-9pt}
			\begin{lstlisting}[language=Python]
				def serve_one(env, res, t_arr, waits):
				with res.request() as req:
				yield req
				st = env.now
				yield env.timeout(random.expovariate(SERVICE_RATE))
				waits.append(st - t_arr)
			\end{lstlisting}
			
		\end{tcolorbox}
		\caption{Representative code fragments from the \emph{batch arrivals} category illustrating three implementation templates: procedural (T0), object-oriented (T1), and functional (T2).}
		\label{fig:template-variants}
	\end{figure}
}

\subsubsection{Python-Based Generation of Instruction-Code Pairs under the CTF}

Stage-I data are generated programmatically. For each selected category, we sample mechanism-specific parameters (e.g., arrival/service rates, capacities, behavioral thresholds, repair rates) under stability-aware ranges, and then instantiate one instruction template and one code template to form an instruction--code pair. Parameter ranges and generation rules follow stability-aware sampling constraints in the generation pipeline. Figure~\ref{fig:CTFP} describes this data construction process.

\begin{figure}[t]
	\centering
	\resizebox{0.9\linewidth}{!}{%
		\begin{tikzpicture}[
			node distance = 0.6cm and 1.0cm,
			font=\sffamily\small,
			base/.style={
				draw=none,
				rounded corners=2pt,
				align=center,
				blur shadow={shadow blur steps=4, shadow xshift=1pt, shadow yshift=-1pt}
			},
			startstop/.style={
				base,
				fill=teal!80!black,
				text=white,
				minimum height=10mm,
				minimum width=3.2cm,
				font=\sffamily\small\bfseries
			},
			process/.style={
				base,
				fill=white,
				draw=gray!20,
				thick,
				text=black!80,
				minimum height=9mm,
				minimum width=5.0cm,
				path picture={
					\fill[teal!60] (path picture bounding box.south west) rectangle
					($(path picture bounding box.north west)+(0.15,0)$);
				}
			},
			valid/.style={
				base,
				fill=orange!5,
				draw=orange!50,
				dashed, thick,
				text=black!80,
				text width=3.1cm,
				minimum height=11mm
			},
			arrow/.style={
				->,
				>={Latex[length=2mm, width=1.5mm]},
				thick,
				darkgray
			}
			]
			\node (start) [startstop] {CTF Stage I Entry};
			\node (cat)   [process, below=of start] {\ 1. Select Queueing Category (12)};
			\node (param) [process, below=of cat]   {\ 2. Sample System Parameters};
			\node (inst)  [process, below=of param] {\ 3. Select Instruction Template\\(generated by LLM)};
			
			\node (code)   [process, right=of inst] {4. Select Code Template\\(generated by LLM)};
			\node (insert) [process, below=of code]  {\ 5. Insert Parameters into Templates};
			\node (pair)   [startstop, fill=teal!60, below=of insert] {Raw Instruction--Code Pair};
			
			\node (valid)  [valid, right=0.75cm of pair] {Validator:\\Executability \&\\Output Format};
			\node (data)   [startstop, below=0.75cm of pair] {Stage-I SFT Dataset};
			
			\draw[arrow] (start) -- (cat);
			\draw[arrow] (cat) -- (param);
			\draw[arrow] (param) -- (inst);
			
			\draw[arrow] (inst.east) -- (code.west);
			
			\draw[arrow] (code) -- (insert);
			\draw[arrow] (insert) -- (pair);
			
			\draw[arrow] (pair) -- (valid);
			\draw[arrow] (valid.south) |- (data.east);
			
			\path let
			\p1 = (valid.south),
			\p2 = (data.north),
			\n1 = {max(\x1,\x2)},
			\n2 = {(\y1+\y2)/2}
			in
			coordinate (rectRightMid) at (\n1,\n2)
			node[anchor=west, font=\scriptsize, text=teal] at ($(rectRightMid)+(2pt,-0.5)$) {Pass};
		\end{tikzpicture}
	}
	\caption{Stage I data construction pipeline of the CTF.}
	\label{fig:CTFP}
\end{figure}

CTF also provides a quality-control advantage: once a template has been verified for executability, output contract, and mechanism consistency, its instantiated samples inherit these properties. To further reduce risk, we run automatic checks for executability and output-format compliance before including samples in Stage-I training.

\subsubsection{Supervised Fine-Tuning Using the CTF-Constructed Dataset}

Stage~I SFT trains instruction-to-script generation on CTF pairs that satisfy executability and output-contract checks. This stage establishes base capability for full-script queueing simulation generation across categories and implementation styles, with particular emphasis on broad mechanism coverage.

\subsubsection{Supervised Fine-Tuning Using the CTF-Constructed Dataset}

In Stage~I, supervised fine-tuning is performed on CTF-generated instruction--code pairs that satisfy executability and output-contract checks. This stage establishes a foundation for queueing simulation model translation across categories and implementation styles, with particular emphasis on broad mechanism coverage.

\subsection{Stage II: Masked-Completion Data Construction and Understanding-Oriented Fine-Tuning}

In Stage~II, the simulation goal is to strengthen reconstruction of structured event logic by training the model to reconstruct missing functional blocks from partial scripts.

\subsubsection{Motivation of Stage II}

Stage~I may produce executable scripts with mechanism-level logic errors. Masked completion addresses this limitation by forcing recovery of mechanism-critical segments from context, rather than relying only on full-script cues associated with scripts that are merely executable but may still contain mechanism-level logic errors. This stage therefore strengthens the model's ability to reconstruct compositional event logic in simulation programs.

{\linespread{0.8}
	\begin{figure}[t]
		\small
		\centering
		\setlength{\abovecaptionskip}{2pt}
		\setlength{\belowcaptionskip}{0pt}
		\begin{tcolorbox}[
			enhanced,
			width=.96\columnwidth,
			colback=white,
			colframe=black!60,
			boxsep=1pt,left=1pt,right=1pt,top=1pt,bottom=1pt,
			title={\textbf{Masked-Completion Examples for the \emph{Batch Arrivals} Category (Stage II)}}]
			
			\textbf{A. Busy-time accumulation} --- Complete the update of busy-time over $[0,\mathrm{SIM\_TIME}]$.\\[-4pt]
			\begin{lstlisting}[language=Python,basicstyle=\ttfamily\scriptsize,aboveskip=1pt,belowskip=1pt]
				# ... inside service block ...
				s = random.expovariate(SERVICE_RATE); ed = st + s
				# TODO: add busy-time accumulation (overlap with [0, SIM_TIME])
				yield env.timeout(s)
			\end{lstlisting}
			
			\textbf{B. Output format (two-line KPI)} --- Restore printing of 
			``Average waiting time'' and ``Utilization''.\\[-4pt]
			\begin{lstlisting}[language=Python,basicstyle=\ttfamily\scriptsize,aboveskip=1pt,belowskip=1pt]
				# ... end of main script ...
				aw = sum(waits)/len(waits) if waits else 0.0
				util = busy_time / SIM_TIME if SIM_TIME > 0 else 0.0
				# TODO: print two lines: Average waiting time / Utilization
			\end{lstlisting}
			
			\textbf{C. Batch loop (multi-customer arrival)} --- At each batch time, create 
			\emph{BATCH\_SIZE} customers simultaneously.\\[-4pt]
			\begin{lstlisting}[language=Python,basicstyle=\ttfamily\scriptsize,aboveskip=1pt,belowskip=1pt]
				# ... inside arrival source ...
				yield env.timeout(random.expovariate(ARRIVAL_RATE)); t0 = env.now
				# TODO: generate BATCH_SIZE customers at same time
			\end{lstlisting}
			
			\textbf{D. Header parameters} --- Fill in the top-level simulation parameters.\\[-4pt]
			\begin{lstlisting}[language=Python,basicstyle=\ttfamily\scriptsize,aboveskip=1pt,belowskip=1pt]
				# TODO: add simulation parameters here
				# RANDOM_SEED = ...; ARRIVAL_RATE = ...; BATCH_SIZE = ...
				# SERVICE_RATE = ...; SIM_TIME = ...
			\end{lstlisting}
			
		\end{tcolorbox}
		\caption{Representative masked-completion tasks in Stage~II for the \emph{batch arrivals} category.}
		\label{fig:stage2-masks-batch}
	\end{figure}
	\FloatBarrier
}

\subsubsection{Mask Construction for Stage-II Masked-Completion Training}

We define masking rules at the \emph{template level} and apply them programmatically to all Stage~I instances, which therefore enables scalable masked-data construction without per-sample annotation. Mask scopes cover arrival/batch logic, service-resource operations, state updates, routing transitions, behavioral modules, and KPI reporting. Representative masks are shown in Figure~\ref{fig:stage2-masks-batch}. After constructing masked-completion instances, we mix them with the original Stage~I instruction--code pairs for Stage~II training. This mixed curriculum improves mechanism reconstruction while preserving end-to-end generation capability and reducing catastrophic forgetting\endnote{Catastrophic forgetting refers to the phenomenon in which a model loses previously acquired knowledge when trained sequentially on new data or objectives; see \cite{Kirkpatrick2017CF}.}.

\subsection{Stage III: Preference Data Construction and DPO-Based Alignment}

Stage~III uses DPO to align generation preferences with mechanism-faithful simulation quality criteria.

\subsubsection{Motivation of Stage~III}

Stage~II improves local reconstruction but does not explicitly rank competing full scripts for the same instruction. Stage~III introduces preference optimization so mechanism-faithful outputs are preferred over plausible but semantically incorrect alternatives. This stage aims to suppress scripts that may run successfully but still contain mechanism-level logic errors.

\subsubsection{Preference Data Construction for Stage~III}

To construct preference data, we run the Stage~II model on a held-out query set and collect outputs that fail any target criterion as rejected candidates $y^{-}$. Corrected and manually validated scripts are used as chosen candidates $y^{+}$, and the resulting contrastive pairs are used for DPO training. This stage trains the model to jointly prefer executability, output-contract compliance, and mechanism consistency. The complete query-and-validation protocol is reported in Section~\ref{subsec:setup}, and representative examples are provided in Figure~\ref{fig:dpo-example-compact}.

{\linespread{1.0}
	\begin{figure}[t]
		\small
		\centering
		\setlength{\abovecaptionskip}{2pt}
		\setlength{\belowcaptionskip}{0pt}
		\begin{tcolorbox}[
			enhanced,
			width=.96\columnwidth,
			colback=white,
			colframe=black!60,
			boxsep=1pt,left=1pt,right=1pt,top=1pt,bottom=1pt,
			title={\textbf{Compact Preference Pair for \emph{Batch Arrivals} (Stage~III --- DPO)}}]
			
			\textbf{Instruction } \;\;
			\emph{Simulate a batch-arrival single-server queue with $\lambda{=}0.510$, batch size $14$, $\mu{=}0.674$, horizon $t{=}1491$.
				Print exactly two lines (six decimals): ``Average waiting time:'' and ``Utilization:''. Return raw Python code only.}
			
			\vspace{2pt}
			\textbf{Chosen ($y^{+}$) --- key excerpts}\\[2pt]
			\begin{lstlisting}[language=Python,basicstyle=\ttfamily\scriptsize,aboveskip=1pt,belowskip=1pt]
				def batch_arrivals(env, server, waits, busy):
				while True:
				yield env.timeout(random.expovariate(ARRIVAL_RATE))
				t0 = env.now
				for _ in range(BATCH_SIZE):                 # [OK] same-time batch creation
				env.process(customer(env, server, waits, busy, t0))
				# ...
				aw = sum(waits)/len(waits) if waits else 0.0
				util = busy[0] / SIM_TIME if SIM_TIME > 0 else 0.0     # [OK] overlap-clipped utilization
				print(f"Average waiting time: {aw:.6f}")               # [OK] exact two-line format
				print(f"Utilization: {util:.6f}")
			\end{lstlisting}
			
			\vspace{2pt}
			\textbf{Rejected ($y^{-}$) --- key excerpts}\\[2pt]
			\begin{lstlisting}[language=Python,basicstyle=\ttfamily\scriptsize,aboveskip=1pt,belowskip=1pt]
				def batch_arrivals(env, server, waits, busy, up_down):
				yield env.timeout(random.expovariate(ARRIVAL_RATE))
				env.process(customer(env, server, waits, busy, up_down))   # [ERR] single arrivals only
				# ...
				env.process(up_down_proc(env, server, up_down))                # [ERR] irrelevant failure logic
				# ...
				# (format inconsistencies / extra outputs omitted)
			\end{lstlisting}
		\end{tcolorbox}
		\caption{DPO training example with abbreviated code; only key differences are shown for brevity.}
		\label{fig:dpo-example-compact}
	\end{figure}
}

\section{Experimental Design}\label{sec:experiments}

We evaluate the models trained under the proposed simulation-oriented multi-stage framework on a held-out benchmark of queueing tasks. The experiments focus on three evaluation dimensions: executability, output-format compliance, and instruction-mechanism consistency across diverse queueing categories. The following subsections report setup details, evaluation protocol, and stage/category-level results.

\subsection{Experimental Setup}\label{subsec:setup}

We describe models, data, training configurations, and evaluation protocol.

\paragraph{Models and Initialization.} We fine-tune two open-source code language models, Qwen2.5-Coder-7B and DeepSeek-Coder-6.7B, as representative compact code LLMs. Both models are initialized from their publicly released instruction-tuned checkpoints and use identical tokenizer configurations for a fair comparison.

\paragraph{Training Data.} All training data are derived from the multi-stage data construction and fine-tuning framework. Stage~I constructs 7{,}200 validated instruction-code pairs  (about 600 per category on average) for supervised fine-tuning. In Stage~II, the model is trained on a dataset of 20,000 samples, which includes both the newly created masked-completion data and the original Stage~I data. Stage~III constructs 380 human-validated preference pairs for Qwen2.5-Coder-7B and 420 for DeepSeek-Coder-6.7B.

\paragraph{Training Configurations.} All stages are trained using the AdamW optimizer under cosine decay scheduling with a weight decay of 0.1, and a maximum sequence length of 2048 tokens.  Stage~I is trained for three epochs with a learning rate of $2\times10^{-5}$, Stage~II continues for two epochs with a reduced rate of $1\times10^{-5}$, and Stage~III applies DPO for one to two epochs, with the preference strength parameter set to $\beta = 0.2$. Training is performed on a single NVIDIA RTX 4090 GPU (24\,GB memory) with a batch size of 2 and a maximum sequence length of 2048 tokens.

\paragraph{Test Set and Evaluation Metrics.}
Evaluation is conducted on a held-out test pool of 600 tasks across 12 categories, balanced by difficulty level and instruction variant at construction time. The benchmark is designed as a queueing-simulation mechanism benchmark rather than a free-form prompt benchmark. The basic categories cover standard service-resource structures, the intermediate categories cover customer behavior and nonstationary input mechanisms, and the complex categories cover networked transfer, joint-resource coordination, and interruption-rich systems. These groups correspond to representative simulation-modeling difficulties rather than to template buckets alone.
For category-wise reporting, rates are computed on the full category test samples (\(n_c\), 50 per category in our benchmark). Timeout cases, parser violations, and malformed outputs are retained and counted as failures under the corresponding metrics rather than removed from the denominator.
To prevent information leakage, we enforce strict split isolation: the Stage-I/II training pool, the Stage-III query set used for preference construction, and the final test set are instruction-level disjoint.
In addition, template instantiations are generated with non-overlapping parameter seeds across splits. This setting evaluates generalization over held-out task instances and previously unseen instruction-level combinations within the targeted queueing mechanism space, rather than open-ended industrial descriptions.

Each model is evaluated by running its generated code scripts and measuring three criteria:  
\begin{enumerate}
	\item \textbf{executability rate}: which determines whether the generated simulation code runs to completion without runtime errors;
	\item \textbf{output format compliance}: which validates that the program produces the required standardized reporting lines (e.g., average waiting time and utilization) in the correct numerical format;
	\item \textbf{instruction-mechanism consistency}: which assesses whether the operational logic implemented by the generated code faithfully satisfies the simulation specification described in the instruction.
\end{enumerate}
The first two criteria are straightforward to evaluate, as both can be automatically evaluated through code execution and output validation in Python environment. A code script passes the \textbf{executability} check if it runs to completion within the time limit without raising runtime errors, and passes the \textbf{output format} check if it prints the required output lines, such as \texttt{Average waiting time} and \texttt{Utilization}. Because standardized reporting occurs at the end of a successful simulation run, EXE and FMT are expected to be close, although they are not identical when scripts run but violate the required reporting contract.

However, validating instruction-mechanism consistency is challenging as it requires reviewing whether the implemented simulation logic truly matches the intended queueing mechanisms. Accordingly, IMC is used here as a proxy for simulation validity rather than a pure code-generation score. The assessment of IMC is conducted by a large-model-based evaluator, following recent work showing that LLM-based judges can provide reliable semantic evaluation for code-generation tasks \citep{tong2024LLMjudge}.
In our setting, \textbf{DeepSeek} acts as an independent evaluator: for each generated code script, it examines the script together with the corresponding task description and determines whether the code script satisfies the required logic and semantics. A script receives IMC\(=1\) only when the specified queueing mechanisms are implemented completely and correctly; executable but behaviorally invalid scripts are counted as failures. This strict criterion also explains why pre-trained baseline IMC can remain at or near zero even when EXE is nonzero: many baseline outputs are executable but omit or misimplement required simulation mechanisms.
To improve evaluation credibility, we use a structured human spot-check protocol in which sampled cases are stratified across category and complexity groups; disagreement cases are re-reviewed with rule-based adjudication; and finalized disagreements are used to refine parser constraints and evaluator instructions before final scoring.

As part of implementation, the evaluation process is driven by a structured system prompt. The prompt enforces strict rules to ensure that each submission uses the appropriate \texttt{SimPy} primitives, such as \texttt{Environment}, \texttt{Resource}, \texttt{Process}, and \texttt{request}. It further requires that the code implements all mechanisms explicitly specified by the task description, including capacity limits, balking and reneging behaviors, and routing or feedback in networked systems. Code scripts that merely mention relevant keywords without implementing their functional logic are rejected. This design ensures that the consistency metric reflects whether the generated script can reasonably function as a simulation-faithful script. The complete evaluator prompt and parser rules are provided in Figure~\ref{fig:evaluator-prompt-strict}.

{\linespread{0.8}				
	\begin{figure}[t]
		\small
		\centering
		\setlength{\abovecaptionskip}{1pt}
		\setlength{\belowcaptionskip}{1pt}
		\begin{tcolorbox}[
			enhanced,
			width=.95\columnwidth,
			colback=white,
			colframe=black!50,
			boxsep=1pt,left=2pt,right=2pt,top=1pt,bottom=1pt,
			title={\textbf{Evaluation Prompt for \emph{Instruction-mechanism Consistency}}}]
			
			\textbf{SYSTEM\_PROMPT } \;\;
			\emph{You are an expert evaluator for queueing theory and discrete-event simulation code.  
				Task: Based on the ``task description + provided Python/SimPy code,''  
				determine whether the implementation \textbf{follows the intended specification}.}
			
			\vspace{2pt}
			\textbf{Evaluation Principles:}
			\begin{enumerate}
				\item  \textbf{Executable: } The generated script must run from start to finish without raising syntax errors, runtime exceptions, or unresolved references.
				\item \textbf{Structural integrity:}  
				The code must clearly include core \texttt{SimPy} elements such as \texttt{Environment}, \texttt{Resource}, \texttt{Process}, \texttt{request}, and \texttt{timeout}.
				\item \textbf{Precise mechanism matching:}  
				The mechanisms required in the prompt must be correctly implemented in the code logic, not merely mentioned or partially simulated.
				\item \textbf{Logical correctness:}  
				The generated code must be logically sound and fully consistent with the task description, implementing all required processes and mechanisms without missing or irrelevant components.
			\end{enumerate}
			
			\vspace{1pt}
			\textbf{Expected Output (strict JSON format):}
			\begin{lstlisting}[language=Python,basicstyle=\ttfamily\scriptsize,aboveskip=1pt,belowskip=1pt]
				{"label": 1 or 0, "reasons": ["brief reason", "..."]}
			\end{lstlisting}
			
			\vspace{1pt}
			\textbf{Explanation:}
			\begin{itemize}
				\item \texttt{"label": 1} means all structures and mechanisms are complete and logic is correct.
				\item \texttt{"label": 0} means missing key mechanisms or incorrect logic.
				\item No extra natural-language commentary beyond the JSON is allowed.
			\end{itemize}
		\end{tcolorbox}
		\caption{Strict evaluation prompt used by the large-model-assisted evaluator for automatic instruction-consistency assessment.}
		\label{fig:evaluator-prompt-strict}
	\end{figure}
	\FloatBarrier
}

\section{Main Results}

\label{subsec:main-results}

This section reports the performance of the models after training on the held-out test set.
Across the benchmark, models after training are much more likely to produce executable scripts that also preserve the intended queueing mechanisms. For this task, executable code is useful only if it can also serve as a usable queueing simulation script.

\begin{table}[t]
	\centering
	\small
	\begin{tabular}{lcccc}
		\toprule
		\multirow{2}{*}{\textbf{Metric}} 
		& \multicolumn{2}{c}{\textbf{Qwen2.5-Coder-7B}} 
		& \multicolumn{2}{c}{\textbf{DeepSeek-Coder-6.7B}} \\
		\cmidrule(lr){2-3} \cmidrule(lr){4-5}
		& Pre-trained & Fine-tuned & Pre-trained & Fine-tuned \\
		\midrule
		EXE (\%) & 80.4 & \textbf{86.0} & 26.1 & \textbf{75.0} \\
		FMT (\%) & 80.4 & \textbf{86.0} & 9.5 & \textbf{74.8} \\
		IMC (\%) & 0 & \textbf{76.8} & 0 & \textbf{62.3} \\
		\bottomrule
	\end{tabular}
	\caption{Performance comparison on the held-out test set.}
	\label{tab:main-results-overall}
	
\end{table}

As shown in Table~\ref{tab:main-results-overall}, adaptation on the proposed data leads to substantial improvement across all three evaluation metrics. For some models, FMT closely tracks EXE because successful execution is tightly coupled with completion of the fixed reporting block required by the benchmark.
For \textbf{Qwen2.5-Coder-7B}, EXE and FMT both rise from 80.4\% to 86.0\%, while IMC improves from 0\% to 76.8\%. From a simulation viewpoint, this means that many outputs that were previously only executable become usable as queueing simulation scripts with the intended mechanism logic.
\textbf{DeepSeek-Coder-6.7B} exhibits an even larger change in practical usability, with EXE increasing from 26.1\% to 75.0\%, FMT from 9.5\% to 74.8\%, and IMC from 0\% to 62.3\%. For a model that starts from a much weaker baseline on queueing-style simulation tasks, domain adaptation sharply lowers the chance of getting code that runs but implements the wrong mechanism.

These results indicate more than improved executability or output formatting. They show that the workflow produces simulation scripts that are both more usable and closer to the intended mechanism, which is the main requirement in queueing model translation.

\subsection{Category-wise Results}\label{subsec:category}

In this section, we summarize category-level behavior to show which queueing mechanisms are easier to learn and which remain fragile. Tables~\ref{tab:qwen_dpo_temp} and~\ref{tab:deepseek_dpo_temp} report the full per-category results for Qwen2.5-Coder-7B and DeepSeek-Coder-6.7B, respectively. All category-level pass rates are computed on full category test samples, with timeout/parser/output-contract failures retained as failed instances.

{\linespread{1.0}
	\begin{table}[t]
		\centering
		\small
		\renewcommand{\arraystretch}{0.9}
		\setlength{\tabcolsep}{4pt}
		\begin{tabular}{lcccccc}
			\toprule
			\multirow{2}{*}{\textbf{Template}} &
			\multicolumn{3}{c}{\textbf{Pre-trained }} &
			\multicolumn{3}{c}{\textbf{Qwen2.5-Coder-7B (DPO)}} \\
			\cmidrule(lr){2-4} \cmidrule(lr){5-7}
			& EXE & FMT & IMC & EXE & FMT & IMC \\
			\midrule
			\multicolumn{7}{l}{\textit{(A) Basic and Stationary Systems}} \\
			\midrule
			finite\_capacity           & 100.0 & 100.0 & 0.0 & 100.0 & 100.0 & 97.5 \\
			general\_distributions     & 100.0 & 100.0 & 0.0 & 100.0 & 100.0 & 88.5 \\
			multi\_server\_sched\_rules & 100.0 & 100.0 & 0.0 & 100.0 & 100.0 & 87.7 \\
			\midrule
			\multicolumn{7}{l}{\textit{(B) Intermediate Behavioral Extensions}} \\
			\midrule
			balking\_reneging         & 5.0 & 5.0 & 0.0 & 67.5 & 67.5 & 57.5 \\
			batch\_arrivals           & 90.0 & 90.0 & 0.0 & 87.5 & 87.5 & 72.5 \\
			multi\_class\_customers   & 82.5 & 82.5 & 0.0 & 95.0 & 95.0 & 82.5 \\
			piecewise\_arrival        & 90.3 & 90.3 & 0.0 & 93.5 & 93.5 & 62.9 \\
			production\_kanban        & 100.0 & 100.0 & 0.0 & 100.0 & 100.0 & 82.8 \\
			\midrule
			\multicolumn{7}{l}{\textit{(C) Complex Networked or Multi-Mechanism Systems}} \\
			\midrule
			breakdown\_maintenance    & 52.5 & 52.5 & 0.0 & 70.0 & 70.0 & 32.5 \\
			parallel\_two\_resources  & 100.0 & 100.0 & 0.0 & 96.9 & 96.9 & 95.4 \\
			open\_network             & 93.0 & 93.0 & 0.0 & 93.0 & 93.0 & 33.3 \\
			feedback\_network         & 55.0 & 55.0 & 0.0 & 60.0 & 60.0 & 52.5 \\
			\midrule
			\textbf{Average} & \textbf{80.4} & \textbf{80.4} & \textbf{0.0} &
			\textbf{86.0} & \textbf{86.0} & \textbf{76.8} \\
			\bottomrule
		\end{tabular}
		\caption{Performance of \textbf{Qwen2.5-Coder-7B} before and after fine-tuning (DPO stage) across 12 simulation templates. Values are pass rates (\%) computed on full category test samples.}
		\label{tab:qwen_dpo_temp}
	\end{table}
}

Table~\ref{tab:qwen_dpo_temp} summarizes the performance of \textbf{Qwen2.5-Coder-7B} before and after the full three-stage adaptation process across the twelve queueing-system templates. The easiest mechanisms for the model are the basic service-resource structures. After adaptation, \emph{finite\_capacity}, \emph{general\_distributions}, and \emph{multi\_server\_sched\_rules} all achieve very high IMC, which means the model can usually translate standard arrival, service, capacity, and simple scheduling logic into executable \texttt{SimPy} scripts.

The intermediate categories show that the model also learns customer-behavior and input-process extensions reasonably well. Gains for \emph{balking\_reneging}, \emph{batch\_arrivals}, \emph{multi\_class\_customers}, \emph{piecewise\_arrival}, and \emph{production\_kanban} indicate that the workflow supports generation of scripts that preserve queueing rules beyond stationary single-class settings. From a simulation perspective, these categories determine whether the generated script can represent congestion response, bursty demand, differentiated service, or time-varying load.

The hardest mechanisms are the ones with multi-node dependencies and interruption-resume logic. Although \emph{parallel\_two\_resources} becomes highly reliable, \emph{breakdown\_maintenance} and \emph{open\_network} remain much less consistent than the easier categories. Routing transfer, interruption handling, and coupled state updates are still the main barriers to simulation-faithful model translation.

It is also worth noting that a small number of templates such as \emph{parallel\_two\_resources} and \emph{batch\_arrivals} display mild decreases in EXE or FMT after training. This reflects a trade-off in which the workflow places greater emphasis on preserving mechanism logic, so a small amount of executability can be sacrificed for stronger mechanism fidelity.

We next examine the category-wise behavior of \textbf{DeepSeek-Coder-6.7B}. Table~\ref{tab:deepseek_dpo_temp} summarizes its performance before and after adaptation across the same twelve queueing-system templates, so we can see how the workflow changes simulation model generation quality for a much weaker pre-trained starting point. Similar to Qwen2.5-Coder-7B, the gains cover basic, intermediate, and complex queueing categories, and the resulting scripts are much more usable.

{\linespread{1.0}
	\begin{table}[t]
		\centering
		\small
		\renewcommand{\arraystretch}{0.9}
		\begin{tabular}{lcccccc}
			\toprule
			\multirow{3}{*}{\textbf{Template}} &
			\multicolumn{3}{c}{\textbf{Pre-trained DeepSeek}} &
			\multicolumn{3}{c}{\textbf{DeepSeek-Coder-}} \\
			&&&&\multicolumn{3}{c}{\textbf{6.7B (DPO)}} \\
			\cmidrule(lr){2-4} \cmidrule(lr){5-7}
			& EXE & FMT & IMC & EXE & FMT & IMC \\
			\midrule
			\multicolumn{7}{l}{\textit{(A) Basic and Stationary Systems}} \\
			\midrule
			finite\_capacity            & 32.5 & 12.5 & 0.0 & 67.5 & 67.5 & 35.0 \\
			general\_distributions      & 14.8 & 4.9  & 0.0 & 91.8 & 91.8 & 80.3 \\
			multi\_server\_sched\_rules & 40.4 & 21.1 & 0.0 & 70.0 & 70.0 & 47.4 \\
			\midrule
			\multicolumn{7}{l}{\textit{(B) Intermediate Behavioral Extensions}} \\
			\midrule
			balking\_reneging        & 12.5 & 0.0  & 0.0 & 95.0 & 95.0 & 82.5 \\
			batch\_arrivals          & 17.5 & 5.0  & 0.0 & 95.0 & 95.0 & 62.5 \\
			multi\_class\_customers  & 15.0 & 10.0 & 0.0 & 100.0 & 100.0 & 97.5 \\
			piecewise\_arrival       & 35.5 & 8.1  & 0.0 & 72.6 & 72.6 & 40.3 \\
			production\_kanban       & 8.6  & 1.7  & 0.0 & 70.7 & 70.7 & 63.8 \\
			\midrule
			\multicolumn{7}{l}{\textit{(C) Complex Networked or Multi-Mechanism Systems}} \\
			\midrule
			breakdown\_maintenance   & 42.5 & 17.5 & 0.0 & 87.5 & 87.5 & 82.5 \\
			parallel\_two\_resources & 27.7 & 3.1  & 0.0 & 70.0 & 70.0 & 55.4 \\
			open\_network            & 35.1 & 14.0 & 0.0 & 55.0 & 55.0 & 26.3 \\
			feedback\_network        & 30.0 & 20.0 & 0.0 & 97.5 & 97.5 & 57.5 \\
			\midrule
			\textbf{Average} & \textbf{26.2} & \textbf{9.5} & \textbf{0.0} &
			\textbf{75.0} & \textbf{74.8} & \textbf{62.3} \\
			\bottomrule
		\end{tabular}
		\caption{Performance of \textbf{DeepSeek-Coder-6.7B} before and after fine-tuning (DPO stage) across 12 simulation templates. Values are pass rates (\%) computed on full category test samples.}
		\label{tab:deepseek_dpo_temp}
	\end{table}
}

Unlike Qwen, however, DeepSeek exhibits a less even learning profile across queueing mechanisms. Some basic stationary categories are only moderately improved after adaptation, so pre-training characteristics still influence how reliably the model learns standard queueing structures. At the same time, DeepSeek improves sharply on several intermediate and complex mechanisms, including \emph{balking\_reneging}, \emph{multi\_class\_customers}, and \emph{breakdown\_maintenance}. Even from a much weaker starting point, the workflow still produces large gains in simulation model generation quality.

For both Qwen2.5-Coder-7B and DeepSeek-Coder-6.7B, the easier mechanisms are standard service-resource structures, the middle range is customer-behavior rules, and the hardest cases are routing transfer, interruption-resume, and other multi-node state dependencies. This grouping is informative about the simulation problem because it identifies which parts of queueing model translation still carry the most risk.

\section{Ablation and Error Analysis}\label{subsec:ablation}

To examine how each training stage contributes to learning of queueing mechanisms, an ablation study is conducted based on the overall metrics for Qwen and DeepSeek. We track executability, output-format compliance, and instruction-mechanism consistency across training stages. Table~\ref{tab:ablation_overall_summary} summarizes these stage-wise effects, which reports the aggregated performance changes across the three training stages for both models.

\begin{table}[t]
	\centering
	\renewcommand{\arraystretch}{0.9}
	\begin{tabular*}{0.82\linewidth}{@{\extracolsep{\fill}}lccc}
		\toprule
		\textbf{Model \& Stage} & \textbf{EXE} & \textbf{FMT} & \textbf{IMC} \\
		\midrule
		\multicolumn{4}{l}{\textit{Qwen2.5-Coder-7B}} \\
		\midrule
		Stage~I (SFT-1) & 80.7 & 80.7 & 46.2 \\
		Stage~II (SFT-2) & 85.0 & 85.0 & 70.3 \\
		Stage~III (DPO)   & 86.0 & 86.0 & 76.8 \\
		\midrule
		\multicolumn{4}{l}{\textit{DeepSeek-Coder-6.7B}} \\
		\midrule
		Stage~I (SFT-1) & 74.7 & 74.2 & 44.3 \\
		Stage~II (SFT-2) & 67.2 & 66.8 & 62.7 \\
		Stage~III (DPO)   & 75.0 & 74.8 & 62.3 \\
		\bottomrule
	\end{tabular*}
	\caption{Overall ablation results for Qwen and DeepSeek across three training stages.}
	\label{tab:ablation_overall_summary}
\end{table}

As shown in Table~\ref{tab:ablation_overall_summary}, the two models behave differently across training stages. For \textbf{Qwen2.5-Coder-7B}, performance steadily improves at every stage: EXE and FMT rise from 80.7\% to 86.0\%, and IMC grows from 46.2\% to 76.8\%. Each stage supports learning queueing-specific event logic in a different way: Stage~I provides broad mechanism coverage, Stage~II strengthens structured reconstruction, and Stage~III corrects recurrent failure modes.

In contrast, for \textbf{DeepSeek-Coder-6.7B}, the three-stage workflow produces a distinct learning trajectory. During Stage~II, the masked-completion objective significantly boosts IMC, which raises it from 44.3\% to 62.7\%. However, this gain comes with a reduction in EXE and FMT. From a simulation perspective, this indicates that structured program reconstruction is particularly important for learning queueing-specific event logic, but it does not automatically guarantee stronger end-to-end script generation for this model.

In Stage~III, the DPO-based preference optimization helps correct these effects. EXE and FMT recover to 75.0\% and 74.8\%, respectively, which indicates that preference signals help suppress mechanism-level logic errors that are not captured by executability alone. This adjustment slightly lowers IMC, but it yields a more balanced simulation script generator overall.

These results indicate that Qwen and DeepSeek respond differently to the same workflow, with Stage~II being most useful for reconstructing structured simulation programs and Stage~III being most useful for discouraging simulation-invalid but locally plausible outputs.

\subsection{Error Analysis}\label{subsec:EA}

To better understand the limitations that remain after the full three-stage training process, we examine the generated scripts and focus on the queueing mechanisms that the models still struggle with. Several failures remain executable at the code level but invalidate the simulation logic and distort downstream performance measures. The examples below show where these inconsistencies still appear and what they would change in the resulting simulation outputs. Two common issues that appear both in Qwen2.5-Coder-7B and DeepSeek-Coder-6.7B are introduced as follows.

\paragraph{Network Routing and Custom Movement.} One issue often arises when the model needs to generate code for systems that involve multiple service nodes. Although many of the generated scripts can run successfully, they often fail to maintain the intended custom movement of customers across nodes in an open network. Instead of transferring a customer from one node to the next after service completion, the model may generate a new and unrelated arrival at the downstream node. This breaks the required routing structure and leads to network behavior that does not match the task description.

{\linespread{1.0}			
	\begin{figure}[t]
		\small
		\centering
		\setlength{\abovecaptionskip}{2pt}
		\setlength{\belowcaptionskip}{0pt}
		\begin{tcolorbox}[
			enhanced,
			width=.96\columnwidth,
			colback=white,
			colframe=black!60,
			boxsep=1pt,left=1pt,right=1pt,top=1pt,bottom=1pt,
			title={\textbf{Instruction and Erroneous Code Segment (Open Network Routing)}}]
			
			\textbf{Instruction }\;\;
			\emph{Simulate a two-node open queueing network where jobs arrive at Node~1, 
				receive service, then move to Node~2 with probability $p_{12}$. 
				Run to time $t$ and print exactly two lines: ``Average waiting time:'' and ``Utilization:'' 
				(six decimals). Return Python code only.}
			
			\vspace{4pt}
			\textbf{Error Code}\\[2pt]
			\begin{lstlisting}[language=Python,basicstyle=\ttfamily\scriptsize]
				def arrivals(env, n1, n2):
				while True:
				yield env.timeout(random.expovariate(ARRIVAL_RATE))
				env.process(serve(env, n1, env.now))
				# [ERR] incorrectly spawns new arrivals into Node2
				if random.random() < P12:
				env.process(serve(env, n2, env.now))
			\end{lstlisting}
			
		\end{tcolorbox}
		\caption{Representative failure case for network routing semantics.}
		\label{fig:error1}
	\end{figure}			
}

A full code-level example is provided in Figure~\ref{fig:error1}. Figure~\ref{fig:error1} shows that the model does not preserve the intended custom movement of customers between service nodes in an open network. In this code segment, each external arrival at \texttt{Node1} immediately and independently triggers a possible arrival at \texttt{Node2} through a second call to \texttt{serve} with the current simulation time \texttt{env.now}. As a result, customers at \texttt{Node2} do not originate from completed service at \texttt{Node1}, but are instead created as new and unrelated arrivals. This breaks the intended routing structure, because the flow from \texttt{Node1} to \texttt{Node2} should depend on service completion events and the routing probability \(P_{12}\), rather than on the initial external arrival stream. This error distorts flow balance across stations and can bias throughput, queue length, and utilization estimates even though the script remains executable.

\paragraph{Interrupt-Driven Service and Remaining-Time Management.} A second type of issue appears in scenarios that require interruption-based service logic, including systems with breakdown and repair events. Although the model is generally able to construct the server state machine, initiate interruptions, and resume service after the failure is resolved, it does not always maintain a correct record of the remaining service time at the moment of interruption. This mistake does not prevent the script from running, but it alters the simulated service trajectory by allowing a customer to redo portions of service that were already completed before the breakdown.

A full code-level example is provided in Figure~\ref{fig:error-breakdown}. As shown in Figure~\ref{fig:error-breakdown}, the error lies in the handling of the remaining service time after a breakdown interrupt. At the start of service, the code samples a total service requirement and stores it in \texttt{remaining}. Inside the loop, a breakdown correctly triggers an interrupt, and the code records the elapsed busy time during the current attempt. However, after the interruption, the statement \texttt{remaining = max(0.0, remaining)} does not subtract the elapsed time from \texttt{remaining}. The job therefore resumes service with the same total requirement as before the breakdown, as if no work had been completed. This mistake causes each interruption to restart part of the service instead of resuming from the true residual time. As a result, the simulated service durations are inflated and the estimates of waiting time, service completion timing, utilization, and delay distribution become biased.

Overall, these two recurrent error patterns reveal that the model is still weak in several queueing-system simulation tasks. The first issue shows that the model struggles to preserve custom movement across nodes in networked systems, and the second issue indicates unstable handling of state-dependent updates in interrupt-driven service. These cases illustrate why evaluating simulation code requires more than syntax- or execution-based criteria: a script can run and still fail as a valid queueing simulation script.

{\linespread{1.0}
	\begin{figure}[t]
		\small
		\centering
		\setlength{\abovecaptionskip}{0pt}
		\setlength{\belowcaptionskip}{0pt}
		\begin{tcolorbox}[
			enhanced,
			width=.96\columnwidth,
			colback=white,
			colframe=black!60,
			boxsep=1pt,left=1pt,right=1pt,top=1pt,bottom=1pt,
			title={\textbf{Prompt and Erroneous Code Snippet (Breakdown with Interrupt-Driven Service)}}]
			
			\textbf{Prompt}\;\;
			\emph{Simulate a single-server queue with random breakdowns and repairs. 
				Jobs arrive according to a Poisson process, and the server alternates between ``up'' and ``down'' 
				states with exponential mean time between failures (MTBF) and mean time to repair (MTTR). 
				When a breakdown occurs during service, the job should resume with its remaining service time once the server is repaired. 
				Run the simulation to time $t$ and print exactly two lines: ``Average waiting time:'' and ``Utilization:'' (six decimals).}
			
			\vspace{2pt}
			\textbf{Error Code (erroneous excerpt)}\\[2pt]
			\begin{lstlisting}[language=Python,basicstyle=\ttfamily\scriptsize]
				t_start = self.env.now
				remaining = random.expovariate(SERVICE_RATE)
				while remaining > 0:
				self._on = True
				self._t0 = self.env.now
				try:
				yield self.env.timeout(remaining)
				self.busy_time += self.env.now - self._t0
				self._on = False; self._t0 = None
				remaining = 0.0
				except simpy.Interrupt:
				elapsed = self.env.now - self._t0
				if elapsed > 0:
				self.busy_time += elapsed
				self._on = False; self._t0 = None
				yield self.up_event
				remaining = max(0.0, remaining)
			\end{lstlisting}
			
		\end{tcolorbox}
		\caption{Representative failure case for interrupt-driven service with breakdowns.}
		\label{fig:error-breakdown}
	\end{figure}
	\FloatBarrier				
}

\section{Discussion, Limitations, and Conclusion}\label{sec:discussion_conclusion}

From a simulation-methodology perspective, the main use case of the proposed framework is a human-in-the-loop workflow in which an analyst provides a conceptual queueing description, the model generates a \texttt{SimPy} script, the analyst validates mechanism choices, KPI definitions, and assumptions, and the simulation study then proceeds through scenario refinement and experiment design. Under this workflow, the main value of the proposed approach is not to replace model verification, but to provide a more standardized starting point for conceptual-to-executable translation and to support more efficient early-stage script writing.

This framing also clarifies the practical role of local deployment. Compact locally deployable models are relevant for privacy-sensitive operational settings, internal industrial modeling workflows, and educational or practitioner environments with limited infrastructure. In such settings, a locally deployed model may help reduce part of the implementation burden by redirecting some analyst effort from low-level code writing to conceptual validation and experiment design, while still requiring simulation expertise for model checking and interpretation.

\subsection{Limitations}

The current study focuses on \texttt{SimPy}-based queueing systems rather than broader classes of simulation models. IMC is an LLM-assisted proxy for simulation validity and is not equivalent to formal verification of simulation semantics. Workflow-level efficiency gains are inferred from generation quality and still require validation through studies with human simulation analysts. In addition, the benchmark remains structured and does not yet cover highly open-ended industrial problem descriptions. A further limitation is that the present evaluation emphasizes instruction-mechanism consistency and script usability, but does not yet include a dedicated behavioral sanity-check study on KPI trends under parameter perturbations.

\subsection{Conclusion}

This study addresses the challenge of translating queueing concepts into executable simulation models with mechanism fidelity. Across held-out task instances, the proposed workflow improves executability, reporting readiness, and instruction-mechanism consistency, which makes the generated scripts more usable as queueing simulation scripts. The main contribution is a practical step toward more standardized, reproducible, and mechanism-faithful queueing simulation model construction, with LLMs serving as a tool for more reliable simulation model translation. Future research can move in three practical directions:
\begin{enumerate}
	\item Field Validation in Operational Workflows: Conduct empirical studies with human simulation analysts to measure wall-clock productivity gains and identify interactive debugging patterns in real-world planning environments (e.g., hospital triage or call-center staffing).
	\item Enhanced Mechanism Validators: Develop deterministic, non-LLM-based formal verification tools or execution-trace analyzers to definitively evaluate the routing and interruption semantics of generated code.
	\item Extension to Broader DES Domains: Expand the CTF and training pipeline beyond queueing systems to encompass broader discrete-event simulation scenarios, such as supply chain inventory dynamics, continuous--discrete hybrid systems, and complex manufacturing operations with intricate resource contention.
\end{enumerate}

\begingroup \parindent 0pt \parskip 0.0ex \def\enotesize{\normalsize} \theendnotes \endgroup

\bibliographystyle{plainnat}
\bibliography{SimLLM_refer}

\end{document}